\documentclass[12pt]{article} 
\usepackage[margin=1in]{geometry}
\usepackage{booktabs}
\usepackage{graphicx}
\graphicspath{{../../figs/}}
\usepackage{parskip}
\usepackage{multirow}
\usepackage{hyperref}
\usepackage[utf8]{inputenc}
\usepackage[T1]{fontenc}
\usepackage{textcomp}
\usepackage{hyphenat}
\usepackage{subcaption}
\usepackage{float}
\usepackage{tabularx,array}
\newcolumntype{C}{>{\centering\arraybackslash}X}
\usepackage{amsmath}
\usepackage{amssymb}

\usepackage[backend=biber,sorting=none]{biblatex}
\usepackage{url}

\usepackage{array}
\usepackage{orcidlink}
\newcolumntype{P}[1]{>{\raggedright\arraybackslash}p{#1}}

\usepackage[colorinlistoftodos,textsize=small]{todonotes}

\usepackage[labelfont=bf,labelsep=space]{caption}

\newcommand{\orcidicon}[1]{\ifdefined\orcidlink \orcidlink{#1}\, \fi}
\newcommand{\numLLMs}{nine}

\begin{document} 

\title{What Helps Language Models Predict Human Beliefs: Demographics or Prior Stances?}
\author{
Joseph Malone\orcidicon{0009-0001-8100-1851}$^{1,*}$,
Rachith Aiyappa\orcidicon{0000-0002-0802-8994}$^{1}$,
Byunghwee Lee\orcidicon{0000-0003-2369-2022}$^{2}$,\\
Haewoon Kwak\orcidicon{0000-0003-1418-0834}$^{1}$,
Jisun An\orcidicon{0000-0002-4353-8009}$^{1}$,
and Yong-Yeol Ahn\orcidicon{0000-0002-4352-4301}$^{2,\dagger}$\\
\small $^{1}$Center for Complex Networks and Systems Research, Indiana University, Bloomington, IN, USA\\
\small $^{2}$School of Data Science, University of Virginia, Charlottesville, VA 22903, USA\\
\small $^{*}$\texttt{jpmalone@iu.edu}\\
\small $^{\dagger}$\texttt{yyahn@virginia.edu}\\
}
\date{}
\maketitle

\begin{abstract} 
Beliefs shape how people reason, communicate, and behave. Rather than existing in isolation, they exhibit a rich correlational structure---some connected through logical dependencies, others through indirect associations or social processes. As usage of large language models (LLMs) becomes more ubiquitous in our society, LLMs' ability to understand and reason through human beliefs has many implications from privacy issues to personalized persuasion and the potential for stereotyping. 
Yet how LLMs capture this interrelated landscape of beliefs remains unclear. For instance, when predicting someone's beliefs, what information affects the prediction most---who they are (demographics), what else they believe (prior stances), or a combination of both? We address these questions using data from an online debate platform, evaluating the ability of off-the-shelf open-weight LLMs to predict individuals' stance under four conditions: no context, demographics only, prior beliefs only, and both combined. We find that both types of information improve predictions over a blind baseline, with their combination yielding the best performance in most cases. However, the relative value of each varies substantially across belief domains. These findings reveal how current LLMs leverage different types of social information when reasoning about human beliefs, highlighting both their capabilities and limitations.
\end{abstract}

\newpage
\linespread{1.5}\selectfont

\section{Introduction}\label{sec:introduction} 

Beliefs shape how we interpret and navigate the world.
Rather than existing in isolation, they are shaped by external experience, by others, and by their mutual interactions.
These interrelations form complex networks---belief systems---that structure how we make sense of reality~\cite{WhatAreBeliefSystems}.
Within such networks, a shift in one belief can trigger changes in others, while a belief may resist external influence due to interlocking commitments.
Understanding this rich network structure is therefore critical to understanding human belief dynamics.

While some beliefs are connected through logical necessity, other associations can emerge without any apparent logical connection.
One's demographic background and upbringing provide experiences that lead to certain sets of beliefs; associations between beliefs may also spread through social contagion~\cite{LiberalLattes, ObesitySpread,SocialStructureOfGangHomicide, AssociativeDiffusion}.
Once such associations emerge, however spurious they may be, they can be reinforced and hardened through social processes~\cite{OpinionCascadesAndUnpredictability}.

Despite these insights, systematically studying belief systems at scale remains challenging, primarily due to the lack of large-scale data on belief networks.
Recent advances in large language models (LLMs) offer a new computational approach: because LLMs effectively capture psychological and social structures embedded in human-generated text~\cite{kozlowski2019geometry, GenerativeAIForSocialScience}, they may provide a window into how beliefs interrelate.
Studies have used LLMs for related social inference tasks, such as simulating human behavior~\cite{OutOfOneMany, ReplaceHumanParticipants, aher2023usinglargelanguagemodels, horton2023largelanguagemodelssimulated} and predicting public opinion~\cite{chu2023languagemodelstrainedmedia, kim2024aiaugmentedsurveysleveraginglarge, evaluatingLLMsinGeneratingHCI, jiang2022communitylmprobingpartisanworldviews}.
However, these efforts largely center on specific sociopolitical domains, and existing benchmarks focus on stance detection or opinion prediction within narrow contexts~\cite{chu2023languagemodelstrainedmedia, kim2024aiaugmentedsurveysleveraginglarge, jiang2022communitylmprobingpartisanworldviews, ZeroShot}.
They do not systematically evaluate how well LLMs capture the correlational structure of beliefs across diverse domains---an important gap, as LLMs are prone to hallucinations~\cite{Huang_2024, bang-etal-2023-multitask} and biases~\cite{Hartmann2023ThePI,qi2024representationbiaspoliticalsample}, which could amplify errors or reinforce stereotypes when applied broadly.

A recent study introduced a method to fine-tune an LLM encoder to capture this correlational structure in a latent space~\cite{NeuralEmbedding}.
Although this marks an important step toward understanding belief associations, as LLMs become increasingly embedded in society, we also need to understand their \emph{behaviors}~\cite{rahwan2019machine}---how off-the-shelf LLMs, without task-specific fine-tuning, represent the correlational structure of belief systems.

Here, we ask what helps LLMs predict human beliefs: demographic information (who someone is) or prior expressed stances (what else they believe)? Using a benchmark constructed from an online debate platform, we evaluate several open-weight LLMs and find that both types of information improve predictions, but their relative value varies across domains: demographics matter more for identity-linked beliefs like politics and religion, while prior stances matter more for idiosyncratic domains like sports and entertainment. These findings reveal both the capacities and limitations of current LLMs in reasoning about human beliefs, advancing our understanding of how models leverage different types of social information.

\section{Related Work}\label{sec:related work} 

A growing body of work uses LLMs as tools for social measurement, with applications ranging from simulating human behavior~\cite{OutOfOneMany, ReplaceHumanParticipants, aher2023usinglargelanguagemodels, horton2023largelanguagemodelssimulated} and predicting public opinion~\cite{chu2023languagemodelstrainedmedia, kim2024aiaugmentedsurveysleveraginglarge, evaluatingLLMsinGeneratingHCI, jiang2022communitylmprobingpartisanworldviews} to combating misinformation~\cite{CombattingMisinformationUsingLLMs} and improving personalized recommendations~\cite{lyu2024llmrecpersonalizedrecommendationprompting}.
We build on this line of work but focus on evaluating LLMs as social reasoners---measuring how well their predictions align with belief associations observed in humans.
Most prior studies in this area draw on established survey datasets, such as the General Social Survey\footnote{\href{https://gss.norc.org/}{\url{https://gss.norc.org/}} } or American National Election Studies\footnote{\href{https://electionstudies.org/}{\url{https://electionstudies.org/}} }, which focus on political and sociological beliefs.
We leverage the Debate.org~\cite{DDO} (DDO) dataset, which spans diverse categories including religion, science, sports, and entertainment, allowing us to study belief systems more holistically.

Many studies in this space rely on fine-tuning to achieve strong task performance.
For example, \citeauthor{kim2024aiaugmentedsurveysleveraginglarge}~\cite{kim2024aiaugmentedsurveysleveraginglarge} fine-tune Alpaca-7b on nationally representative survey data, and \citeauthor{OutOfOneMany}~\cite{OutOfOneMany} fine-tune GPT-3 using human subgroup backstories to simulate diverse respondents.
Other approaches use embeddings from models like Sentence-BERT~\cite{Reimers2019Sentence} to represent belief interrelations~\cite{NeuralEmbedding}.
However, fine-tuning requires task-specific data and may limit generalization, while also making it difficult to isolate what base models have learned from pretraining alone.

A separate line of work examines zero-shot LLM performance on social reasoning and stance detection~\cite{ziems2024largelanguagemodelstransform, mu2024navigatingpromptcomplexityzeroshot, ZeroShot}.
We extend this direction by comparing zero-shot blind and demographic prompting against few-shot context belief prompting across diverse domains, evaluating what types of information help LLMs predict human beliefs without task-specific fine-tuning.

\section{Methodology}\label{sec:methodology} 

\subsection{Dataset}

We use the Debate.org (DDO) dataset~\cite{DDO}, which contains data spanning from October 2007 to September 2018.
The dataset provides three key types of information: (1) demographic attributes of users (e.g., age, gender, ethnicity, religious ideology), (2) debate topics with arguments from opposing sides, and (3) stance votes (``agree'' or ``disagree'') from users who indicate which side of the debate they support.
We use this information to construct \emph{belief statements} for evaluating LLM prediction capabilities.

\subsection{Extracting Belief Statements from Debates}

Combining a user's stance (``agree'' or ``disagree'') with the debate topic yields the two components necessary to form a belief: a \emph{proposition} and an \emph{attitude}~\cite{StanfordBelief}.
We generate structured belief statements of the form ``I agree/disagree with the following: [Debate Title],'' following the methodology of \citeauthor{NeuralEmbedding}~\cite{NeuralEmbedding}\footnote{\href{https://github.com/ByunghweeLee-IU/Belief-Embedding}{\url{https://github.com/ByunghweeLee-IU/Belief-Embedding}}}.
Table~\ref{tab:example_beliefs} shows example belief statements.

Not all debate titles were suitable for conversion into coherent belief statements.
Many titles were incomplete, used comparative phrasing (e.g., ``Batman or Superman''), or were otherwise difficult to interpret as clear beliefs.
Following \citeauthor{NeuralEmbedding}~\cite{NeuralEmbedding}, we used GPT-4 to filter out titles that could not reasonably be interpreted as human beliefs.
After these steps and additional pre-processing discussed in \S\ref{subsec:preprocessing}, we obtain 119,118 unique belief statements from 5,775 unique users.

\begin{table}[!htbp]
    \centering
    \small
    \renewcommand{\arraystretch}{1.0} 
    \setlength{\tabcolsep}{0pt}
    \begin{tabular}{l}
        \toprule
        \textbf{Example Belief Statements} \\
        \midrule
        I agree with the following: Euthanasia should be legal. \\
        I disagree with the following: Climate change is a hoax. \\
        I agree with the following: Bishops are more valuable than knights in chess. \\
        I disagree with the following: There is a god. \\
        \bottomrule
    \end{tabular}
    \caption{Examples of structured belief statements generated from Debate.org stance-topic pairs}
    \label{tab:example_beliefs}
\end{table}

\subsection{Data Pre-Processing and Chronological Splitting}\label{subsec:preprocessing}

We first filtered the dataset to exclude any users with fewer than five total belief statements.
We then chronologically ordered each user's beliefs to reflect realistic prediction scenarios---using past beliefs to predict future ones.
Next, we split the data into training and test subsets where every user appears in both:
\begin{itemize}
\item\textbf{Training Set:} Also referred to as ``context beliefs'' (i.e., expressed prior beliefs), these are the oldest 80\% of each user's belief statements, provided to the LLM as context during prediction in the relevant experiment settings (see \S\ref{subsec:task_definition}).
\item\textbf{Test Set:} Also referred to as ``test beliefs,'' these are the remaining 20\% of belief statements, used only for evaluation.
For each test belief, the model predicts the user's stance based on varying input information (blind, demographics, context beliefs, or both).
\end{itemize}

To prevent data leakage, we remove any context belief whose cosine similarity with a test belief exceeded 0.8, computed using Sentence-BERT embeddings~\cite{Reimers2019Sentence}.
This threshold was informed by the default parameter of 0.75 in the Sentence-Transformers~\cite{Reimers2019Sentence}, but adjusted to 0.8 to minimize the exclusion of non-duplicate hard negatives.

\begin{figure}[!htbp]
    \centering
    \begin{subfigure}[!htbp]{0.48\textwidth}
        \centering
        \includegraphics[width=\textwidth]{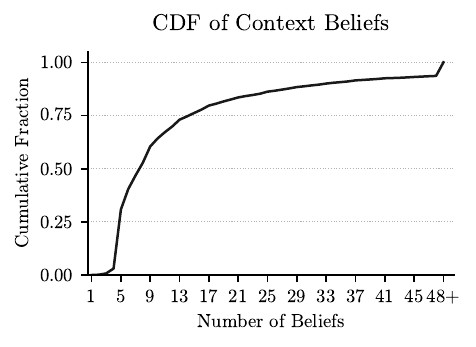}
        \subcaption{Context Beliefs}
        \label{fig:belief_cdf_context}
    \end{subfigure}
    \hfill
    \begin{subfigure}[!htbp]{0.48\textwidth}
        \centering
        \includegraphics[width=\textwidth]{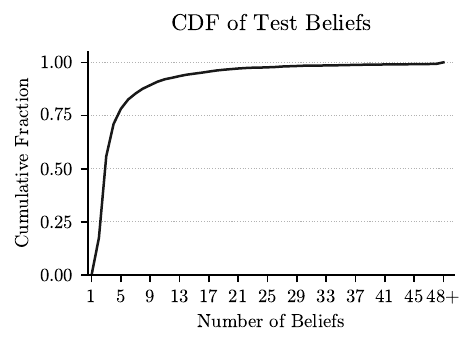}
        \subcaption{Test Beliefs}
        \label{fig:belief_cdf_test}
    \end{subfigure}

    \caption{Cumulative distributions of beliefs per user in the processed DDO dataset: (a) context (training) beliefs and (b) test beliefs. Users with fewer than five total belief statements were excluded}
    \label{fig:belief_distribution_combined}
\end{figure}

Table~\ref{tab:total_beliefs} shows the number of beliefs in each subset and the average number per user after pre-processing.
Figure~\ref{fig:belief_distribution_combined} shows the cumulative distributions of beliefs per user.
Both distributions exhibit heavy-tailed behavior, with a small number of users contributing a disproportionate share of belief statements.

\begin{table}[ht]
    \centering
    \small
    \renewcommand{\arraystretch}{1.0}
    \setlength{\tabcolsep}{6pt}

    \begin{tabular}{l cccc }
        \toprule
        \textbf{Subset} & \textbf{\# Users} & \textbf{\# Beliefs} & \textbf{Avg. Beliefs/User} & \textbf{Std.}\\
        \midrule
        Unfiltered & 40,280 & 192,307 & 4.77  & 21.90\\
        Filtered   & 5,775  & 119,198 & 20.64 & 48.96\\
        Training   & 5,775  & 92,346  & 15.99 & 38.29\\
        Test       & 5,775  & 26,852  & 4.65  & 10.76\\
        \bottomrule
    \end{tabular}
    \caption{The number of users and belief statements contained in each subset, as well as the average number of beliefs per user and standard deviation}
    \label{tab:total_beliefs}
\end{table}

\subsection{Task Definition}\label{subsec:task_definition}

We frame belief prediction as binary classification: given background information about a user, predict their stance (agree/disagree) on a held-out proposition (i.e., a belief statement with the ``I agree/disagree'' prefix removed).
To determine what context is required for accurate prediction, we employ an informational ablation framework, systematically varying the background information provided to the LLM across four conditions: (1) blind (no information about the user, so predictions reflect only population-level priors encoded in the LLM), (2) demographics only, (3) context beliefs only, and (4) both demographics and context beliefs.
We record results under each setting and compare performance across models and belief categories.

\subsection{Models}
Table~\ref{tab:models} shows the \numLLMs~open-weight LLMs used in this study, spanning 3B to 14B parameters.
We focus on open-weight models to ensure reproducibility and because these models can be deployed without specialized infrastructure, making them widely used across diverse applications. Their biases and reasoning capabilities can therefore have broad societal impact, making it important to understand how they reason about human beliefs.
All models were run on an NVIDIA A100 80GB GPU using vLLM\footnote{\url{https://docs.vllm.ai/en/latest/}} for inference.

\begin{table}[!htbp]
    \centering
    \small
    \renewcommand{\arraystretch}{1.0}
    \setlength{\tabcolsep}{6pt}

    \begin{tabular}{l l}
        \toprule
        \textbf{Model} & \textbf{Hugging Face Model ID}\\
        \midrule
        DeepSeek R1 Distill Llama:8b~\cite{deepseekr1} & \href{https://huggingface.co/deepseek-ai/DeepSeek-R1-Distill-Llama-8B/tree/6a6f4aa4197940add57724a7707d069478df56b1}{deepseek-ai/DeepSeek-R1-Distill-Llama-8B}\\
        DeepSeek R1 Distill Qwen:7b~\cite{deepseekr1}  & \href{https://huggingface.co/deepseek-ai/DeepSeek-R1-Distill-Qwen-7B/tree/916b56a44061fd5cd7d6a8fb632557ed4f724f60}{deepseek-ai/DeepSeek-R1-Distill-Qwen-7B}\\
        Gemma 3:4b~\cite{gemma3}                       & \href{https://huggingface.co/google/gemma-3-4b-it/tree/093f9f388b31de276ce2de164bdc2081324b9767}{google/gemma-3-4b-it}\\
        Gemma 3:12b~\cite{gemma3}                      & \href{https://huggingface.co/google/gemma-3-12b-it/tree/96b6f1eccf38110c56df3a15bffe176da04bfd80}{google/gemma-3-12b-it}\\
        Llama 3.1:8b~\cite{llama3}                     & \href{https://huggingface.co/meta-llama/Llama-3.1-8B-Instruct/tree/0e9e39f249a16976918f6564b8830bc894c89659}{meta-llama/Llama-3.1-8B-Instruct}\\
        Llama 3.2:3b~\cite{llama3}                     & \href{https://huggingface.co/meta-llama/Llama-3.2-3B-Instruct/tree/0cb88a4f764b7a12671c53f0838cd831a0843b95}{meta-llama/Llama-3.2-3B-Instruct}\\
        Mistral:7b~\cite{mistral7b}                    & \href{https://huggingface.co/mistralai/Mistral-7B-Instruct-v0.3/tree/0d4b76e1efeb5eb6f6b5e757c79870472e04bd3a}{mistralai/Mistral-7B-Instruct-v0.3}\\
        Phi 4~\cite{phi4}                              & \href{https://huggingface.co/microsoft/phi-4/tree/187ef0342fff0eb3333be9f00389385e95ef0b61}{microsoft/phi-4}\\
        Qwen3:8b~\cite{qwen3}                          & \href{https://huggingface.co/Qwen/Qwen3-8B/tree/b968826d9c46dd6066d109eabc6255188de91218}{Qwen/Qwen3-8B}\\
        \bottomrule
    \end{tabular}
    \caption{Models evaluated in this study}
    \label{tab:models}
\end{table}

\subsection{Prompt Setup}

Chain-of-thought prompting---encouraging LLMs to generate intermediate reasoning steps---improves performance on complex reasoning tasks~\cite{wei2023chainofthoughtpromptingelicitsreasoning}.
LLMs are also highly sensitive to prompt wording~\cite{zhuo-etal-2024-prosa}.
To address both issues, we use DSPy~\cite{DSPy} for its structured input/output design and chain-of-thought module.
DSPy uses \emph{signatures} to clearly structure inputs and outputs, separating what the LLM should do from how it is prompted.
This allows us to organize varying amounts of user background information and extract binary predictions (``agree'' or ``disagree'') while benefiting from chain-of-thought reasoning.

We define model inputs and expected outputs using DSPy's signature interface, which generates consistent system and user messages for each LLM call.
This ensures uniform prompt formatting across models.
Our system message can be found in \S\ref{sec:system_message}, and example user and output messages are provided below.

\subsubsection{Example User Message}
The following shows an example user message for the beliefs + demographics setting.
Other settings use the same layout but exclude the relevant fields.
\begin{quote}
\begin{verbatim}
[[ ## background ## ]]                                                        
description: 23-year old male
education: Some College
ethnicity: Not Saying
income: Less than $25,000
political_ideology: Libertarian                                 
relationship: Married                                               
religious_ideology: Agnostic

[[ ## known_beliefs ## ]]                                                           
[1] «I disagree with the following: Some cryptids are real»
[2] «I agree with the following: x-men movie version would win in a 
battle against avengers movie version»
[3] «I disagree with the following: We are the main focus of God»
[4] «I agree with the following: Humans should be more logic than
faith orientated »
[[ ## proposition ## ]]              
You need a God in order to explain the creation of the universe.

Respond with the corresponding output fields, starting with the field 
`[[ ## reasoning ## ]]`, then `[[ ## prediction ## ]]` (must be 
formatted as a valid Python bool), and then ending with the marker 
for `[[ ## completed ## ]]`
\end{verbatim}
\end{quote}

\subsubsection{Example Output Message}
The following shows an example LLM response to the above user message.
This output format remains the same across all experiment settings.

\begin{quote}
\begin{verbatim}
[[ ## reasoning ## ]]                           
The user expresses atheistic views, disagreeing with
propositions related to God's existence. They also show a preference
for logical reasoning over faith-based arguments (agreeing with ``humans
should be more logic than faith orientated''). Given the user's established
atheism and logic-based
reasoning, they are unlikely to accept this proposition.
[[ ## prediction ## ]]
False
[[ ## completed ## ]]
\end{verbatim}
\end{quote}


\section{Results}\label{sec:results} 

\begin{figure}[!htbp]
    \centering
    \includegraphics[width=1\textwidth]{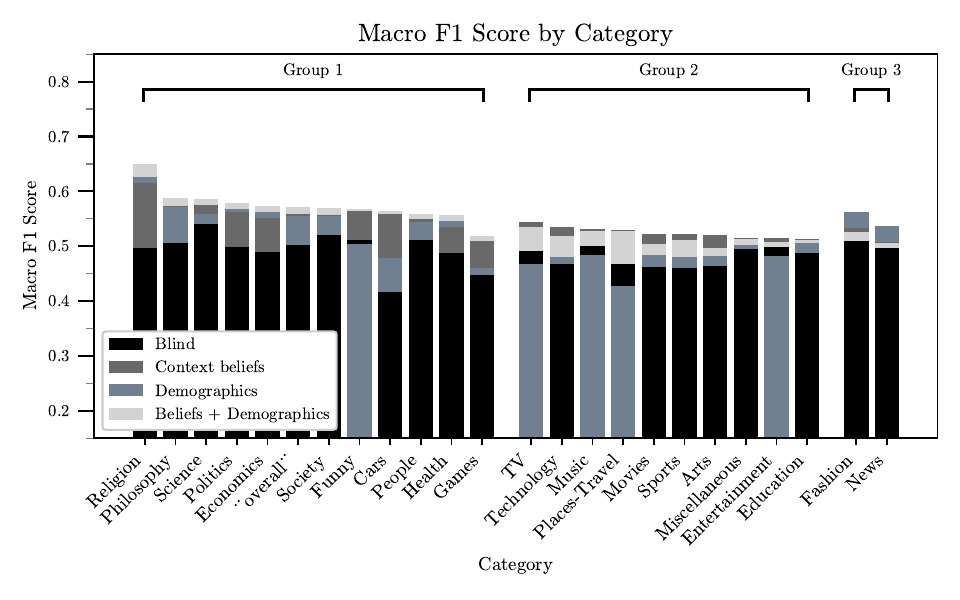}
    \caption{The mean macro F1 scores across models for each prompt setting and belief category separated into three distinct groups based on each category's top-performing experiment setting}
    \label{fig:categorical_stacked_chart}
\end{figure}

\begin{figure}[!htbp]
    \centering
    \includegraphics[width=1\textwidth]{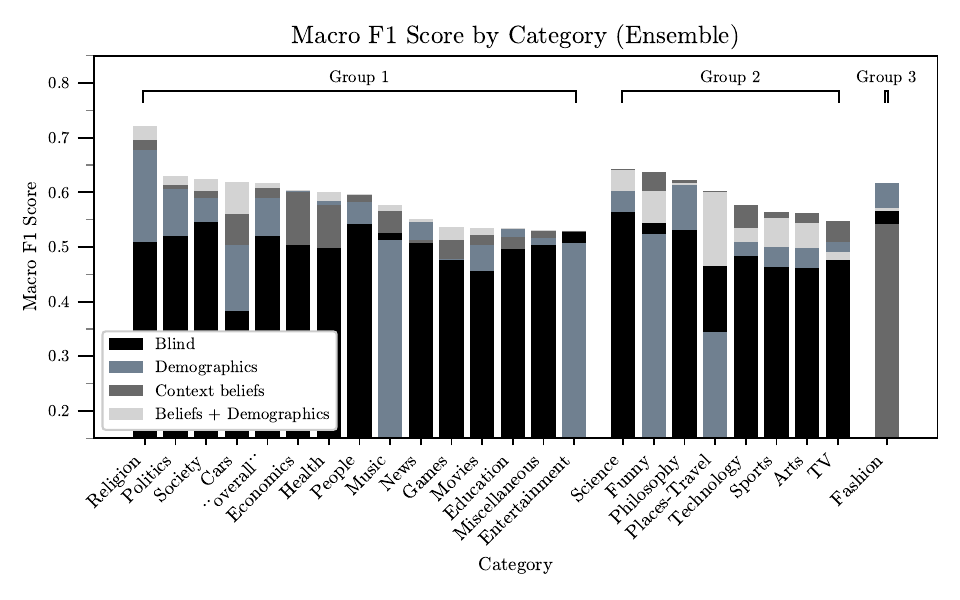}
    \caption{The macro F1 scores when models participate in a majority vote for each prompt setting and belief category separated into three distinct groups based on each category's top-performing experiment setting}
    \label{fig:ensemble_categorical_stacked_chart}
\end{figure}

We aggregate LLM performance using two methods: the mean of individual model scores, which captures typical model behavior, and a majority vote across all LLMs on each prediction, which provides a consensus-based measure.
Figure~\ref{fig:categorical_stacked_chart} shows mean performance across the \numLLMs{} LLMs by belief category and input condition (precise values in Table~\ref{tab:mean_category_f1_results}, \S\ref{sec:additional_results}).
Figure~\ref{fig:ensemble_categorical_stacked_chart} shows majority-vote performance (Table~\ref{tab:ensemble_category_f1_results}, \S\ref{sec:additional_results}).
We organize categories into three groups based on which input condition achieved the highest macro F1 score.

Figures~\ref{fig:categorical_stacked_chart} and~\ref{fig:ensemble_categorical_stacked_chart} reveal three groups based on top-performing input conditions:
\begin{itemize}
    \item \textbf{Group 1 (Demographics + Beliefs):} These categories perform best when provided both demographic attributes and context beliefs.
    Performance typically improves in the order blind $\rightarrow$ demographics $\rightarrow$ context beliefs $\rightarrow$ both, though some categories show demographics outperforming context beliefs or even underperforming the blind baseline.
    \item \textbf{Group 2 (Context Beliefs Only):} These categories perform best with context beliefs alone.
    Adding demographic information to context beliefs degrades performance, though demographics alone still outperforms the blind condition.
    \item \textbf{Group 3 (Demographics Only):} These categories perform best with demographic information alone.
    The ordering of other conditions varies across categories within this group.
\end{itemize}

In general, majority-vote aggregation yields higher macro F1 scores than mean aggregation across categories.
The ranking of input conditions is broadly consistent across both aggregation methods: the combined condition generally performs best, followed by context beliefs, then demographics, with the blind condition performing near chance.

However, in several categories, adding a second information type reduces accuracy compared to using one type alone.
In some cases, demographics even underperforms relative to the blind baseline (e.g., Funny or Music).
At the category level, religious beliefs are predicted most accurately across both aggregation approaches.

\begin{figure}[!htbp]
    \centering
    \includegraphics[width=1\textwidth]{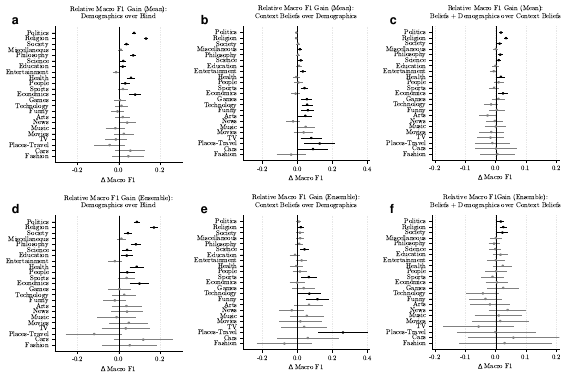}
    \caption{Relative improvement in macro F1 scores across experimental conditions. Columns compare gains from (a, d) adding demographics to the blind baseline; (b, e) using context beliefs in place of demographics; and (c, f) using both demographics and context beliefs. The top row displays average individual model performance (mean), while the bottom row displays the majority vote (ensemble) performance. Error bars indicate 95\% confidence intervals obtained via bootstrapping with 10,000 resamples.
    Statistical significance is indicated by black markers, whereas non-significant estimates are depicted in gray}
    \label{fig:relative_condition_improvement}
\end{figure}

To quantify gains across experiment conditions, Figure~\ref{fig:relative_condition_improvement} presents the relative improvement in macro F1 scores between conditions.
For mean aggregation, we bootstrap observations and compute F1 scores for \emph{each} model on the resampled data, then average across models.
95\% confidence intervals are derived from the distribution of these means over 10,000 resamples.
For majority-vote aggregation, we bootstrap observations and compute F1 from the consensus predictions directly.
95\% confidence intervals are derived from this distribution over 10,000 resamples.

Moving from the blind baseline to demographics (Figures 4a, 4d) yields a consistent positive shift across most categories, particularly in identity-linked domains like Politics and Religion. However, the transition to context beliefs (Figures 4b, 4e) reveals a sharper divergence: while categories like Politics show negligible or negative gains when switching from demographics to context beliefs, idiosyncratic domains such as Places/Travel and Cars exhibit substantial performance jumps. Finally, combining both information sources (Figures 4c, 4f) generally offers only marginal improvements over using context beliefs alone, with confidence intervals frequently crossing zero, indicating that for many categories, the addition of demographics to context beliefs provides little unique predictive value.

These improvement patterns suggest distinct modes of belief formation across domains. For categories like Sports and Entertainment, demographics provide little improvement over the blind baseline, yet context beliefs yield clear gains. This suggests that preferences in these domains are shaped more by idiosyncratic factors---personal experience, upbringing, or local culture---rather than broad demographic categories. In contrast, for Politics, Religion, and Society, demographics alone produce substantial improvements, while context beliefs add comparatively little beyond what demographics already provide. This pattern indicates that beliefs in these domains are more tightly coupled to social identity and group membership.

Table~\ref{tab:model_wise_results} shows individual model performance across input conditions.
Averaged across categories, demographics improve macro F1 over the blind baseline by 6 percentage points, context beliefs by 6.5 points, and the combined condition by 7.8 points.

\begin{table}[!htbp]
    \centering
    \small                             
    \renewcommand{\arraystretch}{1.0}
    \setlength{\tabcolsep}{6pt}
    \begin{tabular}{l cccc }
        \toprule
        & Blind & Dem. & Con. Beliefs & Beliefs + Dem.\\
        \midrule
        DeepSeek R1 Distill Llama:8b & 47.3 & 54.5 & 54.8 & \textbf{56.3}\\
        DeepSeek R1 Distill Qwen:7b  & 50.9 & \textbf{53.4} & 52.2 & 53.1\\
        Gemma 3:4b                   & 51.1 & 55.9 & 55.2 & \textbf{56.6}\\
        Gemma 3:12b                  & 51.7 & 58.9 & 60.7 & \textbf{61.5}\\
        Llama 3.1:8b                 & 47.2 & 53.5 & 51.9 & \textbf{53.6}\\
        Llama 3.2:3b                 & 50.9 & 52.5 & 53.8 & \textbf{55.0}\\
        Mistral:7b                   & 50.1 & 55.4 & 56.8 & \textbf{58.3}\\
        Phi 4:14B                    & 52.5 & 60.0 & 62.0 & \textbf{63.5}\textsuperscript{\textdagger}\\
        Qwen3:8b                     & 52.6 & 59.9 & 61.2 &\textbf{62.4}\\
        \textbf{Combined}            & 50.5 & 56.0 & 56.5 & \textbf{57.8}\\
        \bottomrule
    \end{tabular}
    \caption{The macro F1 Score (\%) of different LLMs under varying input conditions for the belief prediction task.
    ``Combined'' indicates the mean of all models.
    Bold text indicates the best strategy out of each LLM's respective prompt settings. \textsuperscript{\textdagger} indicates the best-performing model
    \label{tab:model_wise_results}}
\end{table}

The combined condition yields the highest macro F1 for 8 of 9 models; DeepSeek R1 Distill Qwen-7B is the exception, peaking under demographics alone.
Improvements from blind to best condition range from +2.5 to +11.0 percentage points (median: +8.2).
Context beliefs outperform demographics in 6 of 9 models, suggesting that expressed prior beliefs are, on average, slightly more informative than demographic attributes.
Phi-4 achieves the highest overall score (63.5\%) in the combined condition.

Overall, LLMs can leverage both demographics and context beliefs to improve belief prediction, but performance varies considerably across categories and models.


\section{Discussion}\label{sec:discussion} 

Our results show that both demographic attributes and prior expressed beliefs tend to improve predictions over the blind baseline, and combining them yields the best performance in most---but not all---belief categories.
However, the optimal input condition varies substantially across domains: some categories benefit most from the combined condition, others from context beliefs alone, and still others from demographics alone. In several cases, adding a second information source actually degrades performance relative to using one source, suggesting that the relationship between input richness and predictive accuracy is not monotonic.
From a machine behavior perspective~\cite{rahwan2019machine}, this pattern indicates that LLMs can leverage contextual cues to personalize inferences beyond broad population tendencies, yet even under the richest conditions, absolute accuracy remains moderate.

Demographic attributes provide high-level social priors---statistical regularities about how people with similar backgrounds tend to reason about different issues.
Prior belief statements, by contrast, offer direct evidence of a user's expressed attitudes and worldview, anchoring predictions in individual-level patterns rather than group-level tendencies.
When combined, we expect that demographics establish a population baseline that context beliefs refine. However, the magnitude and characteristics of this gain varies considerably across belief domains and are not linear. 

The variation in accuracy across belief domains offers further insight into how LLMs capture the structure of human belief systems.
Categories like religion, politics, and science achieve the highest accuracy, as well as the biggest gain via demographic information, likely because these domains are highly salient, composed of sets of coherent beliefs, and tightly coupled with demographics and social identity.
This pattern aligns with the idea of coherent belief systems: some belief domains form densely interconnected networks of mutually reinforcing, coherently interlocked ideas, while others remain loosely organized.
These results suggest that LLMs may be sensitive to this theoretical structure, performing best where beliefs are internally coherent, and struggling more where the associations and coherency among beliefs are not so strong.
Training data composition may amplify these effects, as highly polarized domains tend to produce sharper linguistic contrasts between positions in the text corpora from which models learn.
However, this reliance on demographics is not uniform: for Sports and Entertainment, demographics add little but prior beliefs help, pointing to more idiosyncratic, experience-driven preferences.

Performance also varies across models in ways that reveal something about how they encode social information.
Aggregating predictions via majority vote consistently outperforms averaging individual model scores, suggesting that different models make partially complementary errors.
This complementarity likely stems from differences in architecture, training corpora, and alignment procedures, which lead each model to weight demographic and contextual cues differently.

The moderate overall accuracy highlights inherent limits inference-time belief prediction.
Some of this ceiling reflects noise in the task itself: users' recorded stances may not always reflect their true beliefs, and prior statements are not guaranteed to be consistent with or even relevant to the test proposition.
Additionally, as the dataset spans October 2007 to September 2018, the interrelations captured may not accurately reflect contemporary belief structures.
LLMs thus face both information sparsity and the genuine inconsistency of human reasoning---fundamental challenges that no amount of context can fully resolve.
Model biases further constrain performance: pretrained LLMs absorb the demographic and ideological skews of their training corpora, which can privilege majority viewpoints and produce systematic errors for underrepresented groups.

Without a human baseline, however, it remains unclear whether these models underperform, match, or even exceed human ability to predict others' beliefs from similar information---a question that future work involving human experiments or surveys could address.
Future research could also evaluate frontier closed-source models, explore datasets from different time periods and populations to assess generalizability, and investigate which specific demographic or belief features carry the most predictive weight.

These capabilities and limitations of LLMs carry broader implications.
The ability to infer attitudes from limited cues offers potential value for understanding both model behavior and aspects of human belief systems, but it also raises concerns about privacy, consent, and fairness.
Given the moderate and variable accuracy we observe, applications of belief prediction must be approached with caution: overconfident inferences risk misrepresenting individuals and reinforcing stereotypes.
Responsible deployment requires transparency about uncertainty, clear constraints on use, and careful evaluation of how demographic information shapes model outputs.
Ultimately, belief prediction serves as both a diagnostic tool for characterizing LLM behavior and a testbed for probing how well these models internalize and reproduce patterns of social reasoning.

Future work should extend this framework to state-of-the-art LLMs and establish human baselines to clarify how current models compare to human social reasoning.
Research should also investigate the directionality of belief inference by determining which categories hold the most predictive leverage and identifying asymmetries, such as whether religious beliefs predict political beliefs better than the reverse.
Examining cross-domain reasoning---how models use beliefs in one area to predict attitudes in another---could further reveal whether such patterns reflect genuine social understanding or surface correlations.


\section{Declarations} 
\subsection{Author Contributions}
\textbf{Joseph Malone}: Data curation; Formal analysis; Investigation; Methodology; Software; Visualization; Writing – original draft; Writing – review \& editing

\textbf{Rachith Aiyappa}: Data curation; Methodology, Writing - original draft; Writing – review \& editing

\textbf{Byunghwee Lee}: Data curation; Methodology Writing – review \& editing

\textbf{Haewoon Kwak}: Methodology, Writing – review \& editing

\textbf{Jisun An}: Methodology, Writing – review \& editing

\textbf{Yong-Yeol Ahn}: Conceptualization; Supervision; Methodology; Writing – review \& editing

\subsection{Acknowledgments}
We gratefully acknowledge NVIDIA Corporation for the donation of the GPUs used for this research.

\subsection{Funding}
Byunghwee Lee, Rachith Aiyappa, Jisun An, Haewoon Kwak, and Yong-Yeol Ahn are in part supported by the Air Force Office of Scientific Research under Award No.~FA9550-25-1-0087.
Haewoon Kwak is supported by the Luddy Faculty Fellow Research Grant Program of the Luddy School of Informatics, Computing and Engineering at Indiana University Bloomington.

\subsection{Competing Interests}
The authors have no competing interests to declare that are relevant to the content of this article.

\subsection{Data Availability}
The original DDO dataset is publicly available at: \href{https://esdurmus.github.io/ddo.html}{\url{https://esdurmus.github.io/ddo.html}}.

The processed DDO dataset created by~\citeauthor{NeuralEmbedding}\cite{NeuralEmbedding} is publicly available at: \href{https://github.com/ByunghweeLee-IU/Belief-Embedding}{\url{https://github.com/ByunghweeLee-IU/Belief-Embedding}}

The dataset used in this study that contains additional processing steps, as well as the full replication code, is publicly available at: \href{https://github.com/yy/llm-belief-prediction}{\url{https://github.com/yy/llm-belief-prediction}}.


\pagebreak
\printbibliography{}
\pagebreak
\newpage             
\appendix            
\setcounter{section}{0}
\setcounter{figure}{0} 
\setcounter{table}{0}  

\renewcommand{\thesection}{S\arabic{section}}
\renewcommand{\thefigure}{S\arabic{figure}}
\renewcommand{\thetable}{S\arabic{table}}

\begin{center}
    {\Large \textbf{Supplementary Material}}
\end{center}
\vspace{1em}

\section{Additional Results}\label{sec:additional_results}

\begin{figure}[!htbp]
    \centering
    \includegraphics[width=1\textwidth]{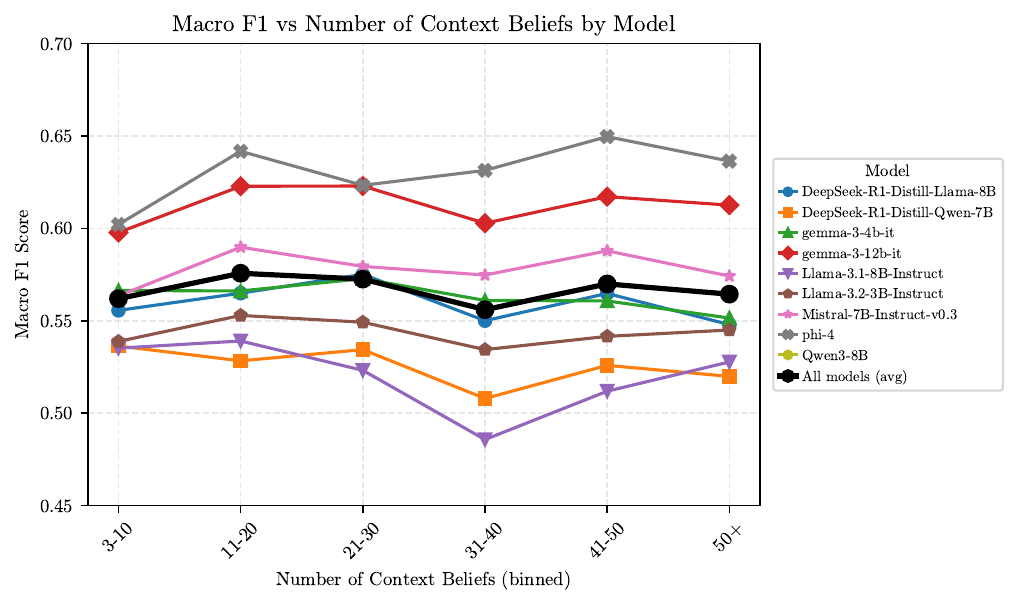}
    \caption{The macro F1 scores for each model when provided varying amounts of context beliefs under the beliefs and demographics experiment setting}
    \label{fig:f1_vs_num_context}
\end{figure}

Figure~\ref{fig:f1_vs_num_context} shows the performance of different models as a function of the number of context beliefs they are provided.
Overall, we see no consistent trend in performance across bins except when models are provided more than 50 context beliefs, at which point performance mostly declines across models.
This decline may reflect limitations in LLMs' ability to integrate and reason over large amounts of information, consistent with other findings that show gaps in LLM capabilities when it comes to understanding longer sequences~\cite{hong2025context}.

Tables \ref{tab:mean_category_f1_results} and \ref{tab:ensemble_category_f1_results} report the macro F1 score of models across all belief categories under each experimental condition and the number of beliefs in each corresponding category.
Overall, we see that richer context improves performance but we also see significant variability at the category level.

\begin{table}[!htbp]
    \centering
    \small
    \renewcommand{\arraystretch}{1.0} 
    \setlength{\tabcolsep}{6pt}

    \begin{tabular}{c l ccccc }
        \toprule
        & & Blind & Dem. & Con. Beliefs & Beliefs + Dem. & Num. Beliefs\\
        \midrule
        \multirow{23}{*}{\rotatebox{90}{Category}}
        & Politics      & 50.1 & 57.5 & 56.8 & \textbf{58.5} & 5815\\
        & Religion      & 50.0 & 63.1 & 62.3 & \textbf{65.7} & 4540\\
        & Society       & 52.4 & 56.0 & 56.3 & \textbf{57.5} & 2771\\
        & Miscellaneous & 49.6 & 50.5 & 51.8 & \textbf{51.8} & 2346\\
        & Philosophy    & 50.7 & 57.7 & 57.9 & \textbf{59.4} & 2047\\
        & Science       & 54.5 & 56.6 & 58.4 & \textbf{59.4} & 1771\\
        & Education     & 49.1 & 50.9 & \textbf{51.7} & 51.6 & 1396\\
        & Entertainment & 50.2 & 48.8 & \textbf{51.8} & 51.1 & 1172\\
        & Health        & 49.0 & 54.9 & 54.2 & \textbf{55.9} & 940\\
        & People        & 51.7 & 54.8 & 55.2 & \textbf{56.2} & 706\\
        & Sports        & 46.4 & 48.4 & \textbf{52.3} & 51.8 & 654\\
        & Economics     & 49.0 & 57.0 & 55.8 & \textbf{58.2} & 646\\
        & Games         & 45.7 & 46.2 & 51.5 & \textbf{52.4} & 375\\
        & Technology    & 47.1 & 48.4 & \textbf{53.8} & 52.3 & 357\\
        & Funny         & 51.4 & 50.7 & 56.6 & \textbf{57.0} & 319\\
        & Arts          & 46.6 & 48.5 & 49.5 & \textbf{50.2} & 242\\
        & News          & 49.8 & \textbf{53.9} & 51.3 & 50.9 & 221\\
        & Music         & 50.4 & 48.7 & 53.3 & \textbf{53.3} & 144\\
        & Movies        & 46.8 & 49.2 & \textbf{52.6} & 51.1 & 127\\
        & TV            & 48.9 & 47.4 & \textbf{55.2} & 53.5 & 104\\
        & Places/Travel & 46.5 & 41.9 & \textbf{54.6} & 53.4 & 58\\
        & Cars          & 41.9 & 47.5 & 56.3 & \textbf{56.4} & 51\\
        & Fashion       & 51.7 & \textbf{56.2} & 52.5 & 53.1 & 50\\
        & \textbf{Combined} & 50.5 & 56.0 & 56.5 & \textbf{57.8} & 26852\\
        \bottomrule
    \end{tabular}
    \caption{The mean macro F1 score of various LLMs on different categories of the belief prediction task.
    Column names correspond to the experiment settings used (blind, demographics, context beliefs, and demographics plus beliefs) and the number of beliefs in each category.
    Bold text indicates the best performer out of each category's respective prompt settings}
    \label{tab:mean_category_f1_results}
\end{table}

\begin{table}[!htbp]
    \centering
    \small
    \renewcommand{\arraystretch}{1.0}
    \setlength{\tabcolsep}{6pt}

    \begin{tabular}{c l ccccc }
        \toprule
        & & Blind & Dem. & Con. Beliefs & Beliefs + Dem. & Num. Beliefs\\
        \midrule
        \multirow{23}{*}{\rotatebox{90}{Category}}
        & Politics      & 51.9 & 60.7 & 61.4 & \textbf{63.0} & 5815\\
        & Religion      & 50.8 & 67.8 & 69.6 & \textbf{72.1} & 4540\\
        & Society       & 54.5 & 58.9 & 60.2 & \textbf{62.4} & 2771\\
        & Miscellaneous & 50.4 & 51.7 & 53.1 & \textbf{53.1} & 2346\\
        & Philosophy    & 53.1 & 61.3 & \textbf{62.2} & 61.7 & 2047\\
        & Science       & 56.3 & 60.3 & \textbf{64.3} & 64.1 & 1771\\
        & Education     & 49.6 & 53.4 & 51.9 & \textbf{53.5} & 1396\\
        & Entertainment & 52.7 & 50.7 & 53.0 & \textbf{53.0} & 1172\\
        & Health        & 49.7 & 58.5 & 57.7 & \textbf{60.0} & 940\\
        & People        & 54.1 & 58.2 & 59.5 & \textbf{59.6} & 706\\
        & Sports        & 46.3 & 50.0 & \textbf{56.4} & 55.3 & 654\\
        & Economics     & 50.4 & 60.0 & 60.1 & \textbf{60.5} & 646\\
        & Games         & 47.6 & 47.9 & 51.3 & \textbf{53.7} & 375\\
        & Technology    & 48.4 & 51.0 & \textbf{57.7} & 53.4 & 357\\
        & Funny         & 54.4 & 52.4 & \textbf{63.7} & 60.3 & 319\\
        & Arts          & 46.1 & 49.7 & \textbf{56.3} & 54.4 & 242\\
        & News          & 50.7 & 54.6 & 51.3 & \textbf{55.2} & 221\\
        & Music         & 52.6 & 51.4 & 56.6 & \textbf{57.6} & 144\\
        & Movies        & 45.6 & 50.3 & 52.1 & \textbf{53.5} & 127\\
        & TV            & 47.6 & 50.9 & \textbf{54.5} & 49.1 & 104\\
        & Places/Travel & 46.5 & 34.5 & \textbf{60.3} & 60.2 & 58\\
        & Cars          & 38.4 & 50.3 & 56.0 & \textbf{61.8} & 51\\
        & Fashion       & 56.6 & 61.8 & 54.2 & \textbf{57.2} & 50\\
        & \textbf{Combined} & 52.0 & 59.0 & 60.7 & \textbf{61.7} & 26852\\
        \bottomrule
    \end{tabular}
    \caption{The macro F1 score of various LLMs using a majority vote on different categories of the belief prediction task.
    Column names correspond to the experiment settings used (blind, demographics, context beliefs, and demographics plus beliefs) and the number of beliefs in each category.
    Bold text indicates the best performer out of each category's respective prompt settings}
    \label{tab:ensemble_category_f1_results}
\end{table}

Similarly, \ref{tab:mean_category_acc_results} and \ref{tab:ensemble_category_acc_results} report model accuracies across all belief categories under each experimental condition.
The number of beliefs in each category is identical to Tables \ref{tab:mean_category_f1_results} and \ref{tab:ensemble_category_f1_results}.

\begin{table}[!htbp]
    \centering
    \small
    \renewcommand{\arraystretch}{1.0}
    \setlength{\tabcolsep}{6pt}

    \begin{tabular}{c l cccc }
        \toprule
        & & Blind & Dem. & Con. Beliefs & Beliefs + Dem.\\
        \midrule
        \multirow{23}{*}{\rotatebox{90}{Category}}
        & Politics      & 51.3 $\pm$ 0.6 & 58.0 $\pm$ 0.7 & 57.1 $\pm$ 0.6 & \textbf{58.8 $\pm$ 0.7}\\
        & Religion      & 51.8 $\pm$ 0.7 & 64.0 $\pm$ 0.7 & 62.8 $\pm$ 0.7 & \textbf{66.0 $\pm$ 0.7}\\
        & Society       & 53.2 $\pm$ 0.9 & 56.6 $\pm$ 0.9 & 56.6 $\pm$ 0.9 & \textbf{57.9 $\pm$ 0.9}\\
        & Miscellaneous & 50.4 $\pm$ 1.0 & 51.3 $\pm$ 1.1 & \textbf{52.3 $\pm$ 1.1} & 52.2 $\pm$ 1.0\\
        & Philosophy    & 52.0 $\pm$ 1.1 & 58.4 $\pm$ 1.1 & 58.2 $\pm$ 1.0 & \textbf{59.7 $\pm$ 1.1}\\
        & Science       & 54.9 $\pm$ 1.2 & 57.3 $\pm$ 1.2 & 58.8 $\pm$ 1.1 & \textbf{59.7 $\pm$ 1.2}\\
        & Education     & 50.1 $\pm$ 1.3 & 51.8 $\pm$ 1.3 & \textbf{52.3 $\pm$ 1.3} & 52.1 $\pm$ 1.3\\
        & Entertainment & 51.3 $\pm$ 1.5 & 49.9 $\pm$ 1.5 & \textbf{52.6 $\pm$ 1.5} & 51.5 $\pm$ 1.5\\
        & Health        & 50.4 $\pm$ 1.6 & 55.7 $\pm$ 1.6 & 54.7 $\pm$ 1.6 & \textbf{56.3 $\pm$ 1.6}\\
        & People        & 52.6 $\pm$ 1.8 & 55.6 $\pm$ 1.8 & 55.7 $\pm$ 1.9 & \textbf{56.5 $\pm$ 1.8}\\
        & Sports        & 47.8 $\pm$ 1.9 & 50.0 $\pm$ 1.9 & \textbf{54.1 $\pm$ 1.9} & 52.7 $\pm$ 2.0\\
        & Economics     & 50.1 $\pm$ 1.9 & 57.5 $\pm$ 1.9 & 56.1 $\pm$ 1.9 & \textbf{58.5 $\pm$ 1.8}\\
        & Games         & 47.2 $\pm$ 2.5 & 47.9 $\pm$ 2.6 & 52.7 $\pm$ 2.6 & \textbf{53.0 $\pm$ 2.5}\\
        & Technology    & 48.7 $\pm$ 2.6 & 49.6 $\pm$ 2.7 & \textbf{54.4 $\pm$ 2.6} & 52.7 $\pm$ 2.6\\
        & Funny         & 53.8 $\pm$ 2.8 & 53.6 $\pm$ 2.7 & \textbf{59.0 $\pm$ 2.8} & 58.3 $\pm$ 2.8\\
        & Arts          & 47.9 $\pm$ 3.1 & 50.2 $\pm$ 3.1 & \textbf{54.3 $\pm$ 3.2} & 51.0 $\pm$ 3.3\\
        & News          & 50.9 $\pm$ 3.5 & \textbf{54.5 $\pm$ 3.3} & 52.1 $\pm$ 3.3 & 51.3 $\pm$ 3.2\\
        & Music         & 51.6 $\pm$ 4.0 & 49.8 $\pm$ 4.2 & \textbf{54.3 $\pm$ 4.0} & 53.7 $\pm$ 4.2\\
        & Movies        & 48.1 $\pm$ 4.5 & 50.3 $\pm$ 4.4 & \textbf{53.6 $\pm$ 4.3} & 51.6 $\pm$ 4.3\\
        & TV            & 49.9 $\pm$ 4.8 & 48.8 $\pm$ 4.9 & \textbf{55.8 $\pm$ 4.7} & 53.8 $\pm$ 4.8\\
        & Places/Travel & 47.7 $\pm$ 6.6 & 43.1 $\pm$ 6.5 & \textbf{55.2 $\pm$ 6.2} & 54.2 $\pm$ 6.5\\
        & Cars          & 44.7 $\pm$ 6.9 & 49.2 $\pm$ 6.8 & \textbf{59.9 $\pm$ 6.9} & 58.2 $\pm$ 6.7\\
        & Fashion       & 54.0 $\pm$ 7.1 & \textbf{58.9 $\pm$ 7.0} & 54.4 $\pm$ 6.9 & 54.7 $\pm$ 6.9\\
        & \textbf{Combined} & 51.5 $\pm$ 0.3 & 56.7 $\pm$ 0.3 & 57.0 $\pm$ 0.3 & \textbf{58.1 $\pm$ 0.3}\\
        \bottomrule
    \end{tabular}
    \caption{The mean accuracy ($\pm$ standard deviation) of various LLMs on different categories of the belief prediction task.
    Column names correspond to the experiment settings used (blind, demographics, context beliefs, and demographics plus beliefs).
    Bold text indicates the best performer out of each category's respective prompt settings.}
    \label{tab:mean_category_acc_results}
\end{table}
\begin{table}[!htbp]
    \centering
    \small
    \renewcommand{\arraystretch}{1.0}
    \setlength{\tabcolsep}{6pt}

    \begin{tabular}{c l cccc }
        \toprule
        & & Blind & Dem. & Con. Beliefs & Beliefs + Dem.\\
        \midrule
        \multirow{23}{*}{\rotatebox{90}{Category}}
        & Politics      & 52.0 $\pm$ 0.7 & 60.7 $\pm$ 0.6 & 61.4 $\pm$ 0.6 & \textbf{63.0 $\pm$ 0.6}\\
        & Religion      & 51.9 $\pm$ 0.8 & 68.3 $\pm$ 0.7 & 69.7 $\pm$ 0.7 & \textbf{72.1 $\pm$ 0.6}\\
        & Society       & 54.6 $\pm$ 0.9 & 58.9 $\pm$ 0.9 & 60.2 $\pm$ 0.9 & \textbf{62.5 $\pm$ 1.0}\\
        & Miscellaneous & 50.4 $\pm$ 1.1 & 51.7 $\pm$ 1.2 & 53.1 $\pm$ 1.0 & \textbf{53.3 $\pm$ 1.1}\\
        & Philosophy    & 53.3 $\pm$ 1.2 & 61.5 $\pm$ 1.1 & \textbf{62.2 $\pm$ 1.1} & 61.8 $\pm$ 1.0\\
        & Science       & 56.4 $\pm$ 1.1 & 60.3 $\pm$ 1.0 & \textbf{64.3 $\pm$ 1.3} & 64.1 $\pm$ 1.1\\
        & Education     & 49.6 $\pm$ 1.4 & 53.4 $\pm$ 1.4 & 51.9 $\pm$ 1.4 & \textbf{53.6 $\pm$ 1.4}\\
        & Entertainment & 52.6 $\pm$ 1.3 & 50.8 $\pm$ 1.6 & \textbf{53.4 $\pm$ 1.3} & 53.0 $\pm$ 1.4\\
        & Health        & 50.0 $\pm$ 1.7 & 58.4 $\pm$ 1.6 & 57.7 $\pm$ 1.4 & \textbf{60.0 $\pm$ 1.7}\\
        & People        & 54.2 $\pm$ 1.8 & 58.5 $\pm$ 1.9 & \textbf{59.8 $\pm$ 1.9} & 59.6 $\pm$ 1.9\\
        & Sports        & 46.3 $\pm$ 2.2 & 50.3 $\pm$ 2.0 & \textbf{58.4 $\pm$ 2.1} & 55.8 $\pm$ 2.3\\
        & Economics     & 50.5 $\pm$ 2.0 & 60.4 $\pm$ 1.8 & 60.2 $\pm$ 1.9 & \textbf{60.7 $\pm$ 1.6}\\
        & Games         & 47.7 $\pm$ 2.5 & 48.0 $\pm$ 2.8 & 53.1 $\pm$ 2.7 & \textbf{53.9 $\pm$ 2.3}\\
        & Technology    & 48.5 $\pm$ 2.7 & 51.0 $\pm$ 2.3 & \textbf{58.0 $\pm$ 2.6} & 53.5 $\pm$ 2.5\\
        & Funny         & 56.4 $\pm$ 2.7 & 54.9 $\pm$ 2.9 & \textbf{66.1 $\pm$ 2.8} & 61.1 $\pm$ 2.8\\
        & Arts          & 46.3 $\pm$ 3.0 & 50.4 $\pm$ 3.2 & \textbf{57.4 $\pm$ 3.0} & 54.5 $\pm$ 3.3\\
        & News          & 50.7 $\pm$ 3.1 & 54.8 $\pm$ 3.6 & 51.6 $\pm$ 3.1 & \textbf{55.2 $\pm$ 3.3}\\
        & Music         & 52.8 $\pm$ 4.6 & 51.4 $\pm$ 4.1 & 56.9 $\pm$ 3.7 & \textbf{57.6 $\pm$ 4.3}\\
        & Movies        & 45.7 $\pm$ 4.9 & 50.4 $\pm$ 4.2 & \textbf{53.5 $\pm$ 3.9} & 53.5 $\pm$ 4.6\\
        & TV            & 48.1 $\pm$ 5.0 & 51.0 $\pm$ 4.6 & \textbf{54.8 $\pm$ 4.4} & 50.0 $\pm$ 5.0\\
        & Places/Travel & 46.6 $\pm$ 6.6 & 34.5 $\pm$ 6.9 & \textbf{60.3 $\pm$ 5.2} & 60.3 $\pm$ 6.6\\
        & Cars          & 39.2 $\pm$ 7.4 & 51.0 $\pm$ 7.1 & \textbf{62.7 $\pm$ 6.7} & 62.7 $\pm$ 6.5\\
        & Fashion       & 58.0 $\pm$ 7.6 & \textbf{64.0 $\pm$ 6.8} & 56.0 $\pm$ 8.1 & 58.0 $\pm$ 6.8\\
        & \textbf{Combined} & 52.0 $\pm$ 0.3 & 59.1 $\pm$ 0.3 & 60.8 $\pm$ 0.3 & \textbf{61.7 $\pm$ 0.3}\\
        \bottomrule
    \end{tabular}
    \caption{The accuracy ($\pm$ standard deviation) of various LLMs using a majority vote on different categories of the belief prediction task.
    Column names correspond to the experiment settings used (blind, demographics, context beliefs, and demographics plus beliefs).
    Bold text indicates the best performer out of each category's respective prompt settings.}
    \label{tab:ensemble_category_acc_results}
\end{table}

\section{System Message}\label{sec:system_message}
This system message is used in the beliefs\hyp{}demographics experiment setting.
Other experiment settings remain unchanged from this setup apart from the exclusion of the specified input fields.
\begin{quote}
\begin{verbatim}
Your input fields are:
1. `background` (str): Demographic information about the user
2. `known_beliefs` (str): Known beliefs of the user
3. `proposition` (str): A proposition we don't know a user's attitude
toward

Your output fields are:
1. `reasoning` (str)
2. `prediction` (bool): True if a user is likely to agree with the
proposition
All interactions will be structured in the following way, with the
appropriate values filled in.
[[ ## background ## ]]
{background}
[[ ## known_beliefs ## ]]
{known_beliefs}
[[ ## proposition ## ]]
{proposition}
[[ ## reasoning ## ]]
{reasoning}
[[ ## prediction ## ]]
{prediction}    # note: the value you produce must be True or False
[[ ## completed ## ]]

In adhering to this structure, your objective is:
        Predict the user's attitude toward the proposition
\end{verbatim}
\end{quote}
\end{document}